\RequirePackage{fix-cm}
\documentclass[smallextended]{svjour3}       
\smartqed  
\usepackage[utf8]{inputenc} 
\usepackage[T1]{fontenc}    
\usepackage{hyperref}       
\usepackage{url}            
\usepackage{booktabs}       
\usepackage{amsfonts}       
\usepackage{nicefrac}       
\usepackage{microtype}      
\usepackage{amsmath}
\usepackage{mathtools}
\usepackage{tikz}
\usepackage{pgfplots}
\usepackage{blochsphere}
\usepackage{graphicx}
\usepackage{verbatim}
\usepackage{subcaption}
\usepackage{tikz}
\pgfplotsset{
  compat=newest,
  xlabel near ticks,
  ylabel near ticks
}
\DeclarePairedDelimiter\bra{\langle}{\rvert}
\DeclarePairedDelimiter\ket{\lvert}{\rangle}
\DeclarePairedDelimiterX\braket[2]{\langle}{\rangle}{#1 \delimsize\vert #2}

\title{Defining Quantum Neural Networks via Quantum Time Evolution}

 \author{Aditya~Dendukuri$^{1}$ \and %
          Blake Keeling$^{1}$\and%
         ~Arash~Fereidouni$^{2,3}$\and%
         ~Joshua~Burbridge$^{1}$\and%
         ~Khoa~Luu$^{1}$\and %
        ~Hugh~Churchill$^{2,3}$ \\
 \institute{$^{1}$Computer Vision and Image Understanding Lab, Computer Science and Computer Engineering Department, $^{2}$Department of Physics, \emph{and} $^{3}$Institute for Nanoscience and Engineering, University of Arkansas, Fayetteville, AR, 72701, Emails: \textit{\{adenduku, bjkeeling afereido\}@email.uark.edu, \{khoaluu, hchurch\}@uark.edu}}}

\begin{document}

\maketitle

\begin{abstract}
This work presents a parameter encoding scheme for defining and training neural networks in quantum information based on time evolution of quantum spaces. Classical neural network algorithms are computationally expensive. For example, in image classification, representing an image pixel by pixel using classical information requires an enormous amount of computational memory resources. Hence, exploring methods to represent images in a different paradigm of information is important. Quantum Neural Networks (QNNs) have been explored for over 20 years. The current forefront work based on Variational Quantum Circuits is specifically defined for the Continuous Variable (CV) Model of quantum computers. In this work, a model is proposed which is defined at a more fundamental level and hence can be inherited by any variants of quantum computing models. This work also presents a quantum backpropagation algorithm to train our QNN model and validate this algorithm on the MNIST dataset on a quantum computer simulation.
\end{abstract}


\section{Introduction}

Deep Learning is a highly successful field of computer science, playing an integral role in cutting-edge technology like pattern recognition and self-driving cars. With recent developments in deep learning, fields like Computer Vision have been revolutionized with significant advancements in pattern recognition, classification, etc. In this work, we focus on the relevance of the quantum framework of information, namely \textit{Quantum Computing}, to image classification using deep learning. Image classification is currently one of the most rapidly-changing research areas in deep learning. Representing image pixel by pixel using classical information requires enormous amounts of computational resources. Hence, exploring  methods to represent images in a different paradigm is important. Deep Neural Networks in Quantum Information or Quantum Neural Networks (QNNs) have been getting a lot of attention in the past couple of years due to recent advancements in quantum computing. One of the most popular methods of QNN's is the Variational Quantum Circuit. Variational Quantum Circuits however have a disadvantage. The frontier models of variational quantum circuits  are only limited to the Continous Variable Paradigm of quantum computers. Hence we need to build the quantum machine learning theory around more fundamental concepts of Quantum Theory which can universally apply to any quantum computing paradigm. In this work, one of the main focal points is to develop the notion of Variational Quantum Circuits in terms of fundamental concepts in quantum theory, specifically Hamiltonian operators and time evolution of quantum states. Our second focus is to test experimentally the performance of variational quantum circuits for the MNIST handwritten digits database. Our contributions can be compiled as follows.\\
\textbf{Contributions: }
This work defines a Quantum Neural Network (QNN) model as a \textit{Hamiltonian} dictating the evolution of quantum states. This work also includes derivation of expression for quantum squared loss and a \textit{Quantum Backpropogation algorithm} was developed based on the definition. This work then demonstrates how this fundamental definition can translate to a specific quantum system by experimentally testing our QNN model by training it to identify handwritten digits on MNIST database in a Quantum Computer Simulation library QuEST \cite{Jones_2018_arxiv}. The proposed method is able to obtain 64\% accuracy for MNIST on a Quantum Neural Network which is the highest accuracy for a large scale data-set to date. 

\subsection{Quantum Computation}

\textit{Quantum Computing} defines a non-deterministic approach to represent classical information using ideas from Quantum Theory. The idea of quantum information was first introduced in 1980 by Paul Benioff \cite{beinoff1980}. In the same year, Yuri Manin proposed a quantum computer in his textbook ``Computable and Uncomputable" \cite{manin1980}. In the year 1982, the field was formalized and made popular by Richard Feynman in his paper about simulating physics in computers \cite{feynman1982}. David Deutsch further advanced the field by formulating a Quantum Turing Machine \cite{deutsch1985}. Since then, there have been a number of algorithms developed such as the Grover's search algorithm \cite{grover1996} and Shor's factoring algorithm \cite{shor1999}. Shor's factoring algorithm is particularly significant since it demonstrated exponential speedup in factoring a large  number. Since, modern day encryption techniques operate using huge numbers, the ideas sprouting from the Quantum Information are already implying effect on our world. The core idea of quantum computing is a \textit{qubit}: the quantum analogue of a classical computer bit. A classical bit is capable of storing a determined value (0 or 1). A qubit (say $\psi$), can be represented using a superposition of both 0 and 1,
\begin{equation} \label{eqn:psi}
    \ket{\psi} = \alpha\ket{0} + \beta\ket{1}.
\end{equation}
where, $\ket{0}$ and $\ket{1}$ are vectors $[\begin{smallmatrix} 1 \\ 0  \end{smallmatrix}]$ and $[\begin{smallmatrix} 0 \\ 1  \end{smallmatrix}]$ respectively. $\alpha$ and $\beta$ are the \textit{probability amplitudes}. These probability amplitudes are represented using complex numbers. Hence, getting a real valued probability would mean to take the modulus squared as follows,
\begin{center}
    $P(\psi = \ket{0}) = |\alpha|^2; \, \, \, \, P(\psi = \ket{1}) = |\beta|^2$.
\end{center}

The probabilities are normalized, i.e. they add up to 1: $|\alpha|^2 + |\beta|^2 = 1$.

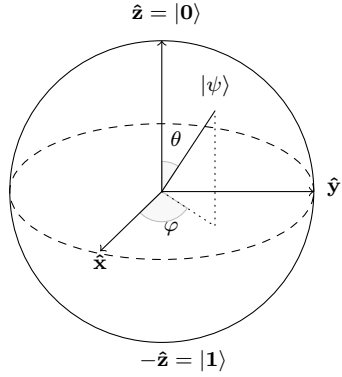
\begin{figure}
  \centering
  \begin{tikzpicture}[line cap=round, line join=round]
  \clip(-3.59,-2.49) rectangle (2.66,2.58);
  \draw [shift={(0,0)}, lightgray, fill, fill opacity=0.1] (0,0) -- (56.7:0.4) arc (56.7:90.:0.4) -- cycle;
  \draw [shift={(0,0)}, lightgray, fill, fill opacity=0.1] (0,0) -- (-135.7:0.4) arc (-135.7:-33.2:0.4) -- cycle;
  \draw(0,0) circle (2cm);
  \draw [rotate around={0.:(0.,0.)},dash pattern=on 3pt off 3pt] (0,0) ellipse (2cm and 0.9cm);
  \draw (0,0)-- (0.70,1.07);
  \draw [->] (0,0) -- (0,2);
  \draw [->] (0,0) -- (-0.81,-0.79);
  \draw [->] (0,0) -- (2,0);
  \draw [dotted] (0.7,1)-- (0.7,-0.46);
  \draw [dotted] (0,0)-- (0.7,-0.46);
  \draw (-0.08,-0.3) node[anchor=north west] {$\varphi$};
  \draw (0.01,0.9) node[anchor=north west] {$\theta$};
  \draw (-1.01,-0.72) node[anchor=north west] {$\mathbf {\hat{x}}$};
  \draw (2.07,0.3) node[anchor=north west] {$\mathbf {\hat{y}}$};
  \draw (-0.5,2.6) node[anchor=north west] {$\mathbf {\hat{z}=|0\rangle}$};
  \draw (-0.4,-2) node[anchor=north west] {$-\mathbf {\hat{z}=|1\rangle}$};
  \draw (0.4,1.65) node[anchor=north west] {$|\psi\rangle$};
  \scriptsize
\end{tikzpicture}
    \caption{Visualization of Rotation and Measurement of quantum state using the Bloch Sphere. $\theta$ and $\varphi$ are the parameters in this case. Tuning them would mean rotating the state $\ket{\psi}$ around the sphere and measuring in z direction means measuring in the computational basis ($\ket{0}, \ket{1}$).}
    \label{fig:bloch}
\end{figure}

\subsubsection{The Bloch Sphere and Quantum Operations}
Given Eqn. \eqref{eqn:psi}, we can expand out the state of a qubit ($\psi$) into a vector as follows:

\begin{equation*}
     \ket{\psi} = \alpha\ket{0} + \beta\ket{1} = \alpha[\begin{smallmatrix} 1 \\ 0  \end{smallmatrix}] + \beta[\begin{smallmatrix} 0 \\ 1  \end{smallmatrix}]=[\begin{smallmatrix} \alpha \\ \beta  \end{smallmatrix}]
\end{equation*}

The \textit{Quantum Logic Gates} that can be applied to qubits come in the form of unitary matrices. Some of these gates are shown in Table \ref{gateexamples}. The function of quantum gates can be visualized in the Bloch spheres (as shown in Figure \ref{fig:bloch}). The fundamental Pauli spin gates ($\Xi :\{ \sigma_x, \sigma_y, \sigma_z, I \}$) form a complete basis. Using this basis we can form any unitary quantum gate (in the form $e^{-i\hat{H}t}$) where $\hat{H} = \sum\limits_{j} a_j \Xi_j$ is the Hamiltonian and t is the time evolution. Any multi qubit quantum gates like the \textit{Controlled NOT} (or CNOT) gate which flips the target qubit if the control qubit is 1 can be decomposed into this basis.

\begin{table}
\begin{tabular}{ |c|c|c|c| } 
 \hline
 \textbf{Name} & \textbf{Dirac Notation} & \textbf{Classical Notation}  & \textbf{Function}\\
 \hline
 PauliX & $\hat{\sigma_x}\ket{\psi}$ & $[\begin{smallmatrix} 0 &1 \\ 1 & 0 \end{smallmatrix}] [\begin{smallmatrix} \alpha \\ \beta  \end{smallmatrix}] = [\begin{smallmatrix} \beta\\ \alpha  \end{smallmatrix}]$ & Rotate in X direction by $\pi$ \\ 
 \hline
 PauliY & $\hat{\sigma_y}\ket{\psi}$ & $[\begin{smallmatrix} 0 &-i \\ i & 0 \end{smallmatrix}] [\begin{smallmatrix} \alpha\\ \beta  \end{smallmatrix}] = i[\begin{smallmatrix} \alpha\\ -\beta  \end{smallmatrix}]$& Rotate in Y direction by $\pi$ \\ 
 \hline
 PauliZ &$\hat{\sigma_z}\ket{\psi}$ & $[\begin{smallmatrix} 1 &0 \\ 0 & -1 \end{smallmatrix}] [\begin{smallmatrix} \alpha\\ \beta  \end{smallmatrix}] = [\begin{smallmatrix} \alpha \\ - \beta  \end{smallmatrix}]$& Rotate in Z direction by $\pi$ \\ 
 \hline
 Hadamard & $\hat{H}\ket{\psi}$ & $\frac{1}{\sqrt{2}}[\begin{smallmatrix} 1 &1 \\ 1 & -1 \end{smallmatrix}] [\begin{smallmatrix} \alpha\\ \beta  \end{smallmatrix}] = \frac{1}{\sqrt{2}}[\begin{smallmatrix} \alpha + \beta\\ \alpha - \beta  \end{smallmatrix}]$& Puts the state in superposition \\ 
 \hline
\end{tabular}
 \caption{Examples and Demonstration of Various Quantum Logic Gates.}
\label{gateexamples}
\end{table}
\section{Neural Networks and Deep Learning}
Classical neural networks are k-partite graphs which represent non-linear transformations. The nodes of the graph may or may not be fully connected. The first ``layer'' represents the input to the network as the nodes. To propagate through the network, we transform the input based on the bond strength between the nodes (weights). Let the input to the network be represented as $\Vec{x}$, and the transformation matrix $\hat{W}$ hold the weights between the connections. The transformation for a layer $L$ can be modelled as:
\begin{equation}
    L_{\hat{W}}(\Vec{x}) = \sigma(\hat{W}\Vec{x} + \Vec{b})
\end{equation}

The bias vector ($\vec{b}$) acts as the intercept of the linear model. The linear model is then passed into a non-linear logistic function to normalize the output. If there are multiple layers, this output will be the input to the next transformation and hence, a neural network can be represented as:

\begin{equation}
    N(\Vec{x}) =  L_{n} \circ L_{n-1} \circ ...... L_{2} \circ L_{1}(\vec{x})
\end{equation}

We can use this function to model a variety of regression and classification problems. The weights and the biases act as tunable parameters which can be adjusted to compute the desired result. This fine-tuning process is termed as ``training'' the neural network. The training is carried out by an algorithm called \textit{backpropogation}.  The backpropogation algorithm minimizes a ``loss'' function which models the accuracy of the network. Mean Squared Error is a common loss function defined as follows: 

\begin{equation}
    \mathcal{L} = (N_{W, b}(\vec{x}) - y(\vec{x}))^2
    \label{loss}
\end{equation}

The training is carried out by updating the weights and biases such that Eqn. \eqref{loss} is minimized as follows:
\begin{equation}
    \Hat{W} = \hat{W} - \eta\frac{\partial\mathcal{L}}{\partial\Hat{W}};\,\,\,\, \,\,\,\vec{b} = \vec{b} - \eta\frac{\partial\mathcal{L}}{\partial\vec{b}}
\end{equation}
The partial derivatives are calculated via the chain rule as it is a sequence of composite functions or layers. Hence, every transformation in the model like the linear and the logistic non-linear transformation of every layer has been fine-tuned to generate a desired output.

\section{Neural Networks in Quantum Information}

\begin{figure}
    \centering
    \includegraphics[scale = 0.35]{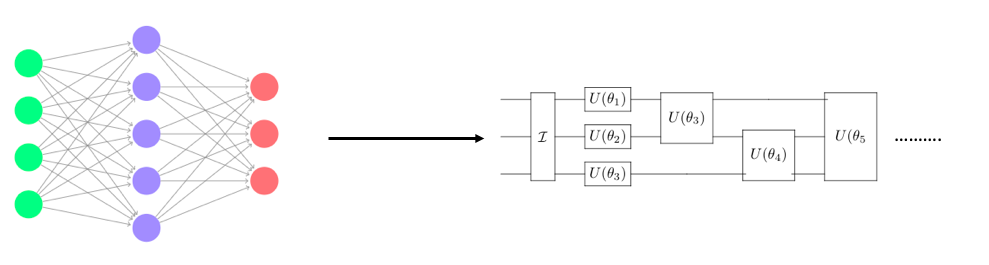}
    \caption{Differentiating between a classical neural network and a variational quantum circuit. Since, input is encoded in a quantum register, there is an exponential reduction in the input space. For the quantum circuit $\mathcal{I}$ is the encode image gate. In addition to that, the rotation gates are naturally computed in a quantum system.}
    \label{fig:schematic}
\end{figure}

Remodelling the theory behind neural networks in quantum information is gaining popularity due to many advancements in quantum computers. The earliest ideas to model a neural network in quantum computing trace back to 1995 paper by Kak \cite{kak1995quantum}. In the same year Menneer and Narayanan proposed a neural network inspired by quantum processes \cite{menneer1995quantum}. Perus \cite{perus1996neuro} suggested the advantage of quantum parallelism being applied to a neural network architecture. The first comprehensive study of a quantum neural network model was conducted by Menneer \cite{meneerthesis}. Altaisky introduced a quantum perceptron model in \cite{altaisky}, but noted that the learning rule for this perceptron did not observe unitarity in general. Gupta and Zia derived  a quantum neural network model from the Deutsch’s model of quantum computational network \cite{guptazia}. A Quantum Neural Network based on qubits was introduced by Kouda \cite{kouda}. Schlud et. al proposed a set of guidelines for developing quantum neural network models\cite{schludquest}. Specifically, a proposal for a generalized quantum neural network should (1) be able to encode some binary string of length N and produce as output some binary string of length M which is closest to N by some distance measure, (2) reflect one or more basic neural computing mechanisms, and (3) be based on quantum effects such as superposition and interference while remaining fully consistent with quantum theory. Quantum machine learning was experimentally tested for two class classification using quantum support vector machine \cite{li2015experimental}. Wiebe et al. introduced the term quantum deep learning in their paper \cite{wiebe2014quantum}. Adachi proposed a quantum neural net model which applies Quantum Annealing \cite{adachi}. A Quantum Boltzmann Machine model was introduced by Amin et al. A quantum recurrent neural network modelled after the Ising Spin Model was proposed in \cite{amin}. Another class of quantum neural networks called variational quantum circuits with tunable parameterized unitary gates implemented in the Continous Variable Model was introduced by Killioran et. al in \cite{killioran}. A Quantum Generative Adversarial Network based on the Variational Circuit Model was introduced by Seth Lloyd in \cite{lloydadv}. 
A more rigorous model focusing on the prospect of near term quantum processors was introduced in \cite{moll2018quantum}.

One difficulty in transplanting quantum computing into machine learning is the general requirement for nonlinear activation functions. A quantum register stores a state vector, which contains the probability amplitudes associated with each possible state. Clearly, this vector is subject to the normalization condition, which means that any operator applied to the system must be unitary. A unitary operator is a square matrix with the property U*U = UU* = I, where U* is the conjugate transpose of U. Thus, the actions on a quantum register are constrained by linear dynamics, which makes quantum computing fundamentally incompatible with the activation function paradigm. 

\subsection{Variational Quantum Circuit}
There are a number of QNN models proposed based on parameterized quantum circuits, also called Variational Quantum Eigensolvers. The circuits composed of parameterized unitary gates that are optimized to produce the desired wave-function. This model can be inferred as a probability distribution similar to output of a softmax function. This is done by repeating and measuring the circuit multiple times and calculating the probability distribution of the basis states. To measure the circuit,observables such as the pauli-z spin matrix ($\sigma_z = [\begin{smallmatrix} 1 , 0 \\ 0 , -1 \end{smallmatrix}]$) can be used to measure in the computational basis as shown in Figure \ref{fig:bloch}). The state of any qubit can be visualized in a Bloch sphere. This kind of circuit with tunable parameters is called a Variational Quantum Circuit as shown in Figure \ref{fig:schematic}). Variational Quantum Circuits belong to a much larger family of \textit{Hybrid} algorithm which require both classical and quantum components \cite{mcclean2016theory}.  

\subsubsection{Advantages and Disadvantages of the Variational Quantum Circuit Model}

There are a number of potential advantages of a Variational Quantum Circuits over  classical deep models as follows:
\begin{itemize}
    \item The first motivation is the notion of quantum parallelism. Since a state can be modeled as it is holding both possibilities, any operation potentially could naturally compute both probabilities at once. 
    \item The natural ability of quantum gates to represent a rotation operator. There exists unitary quantum gates which represent rotations around the Bloch sphere. 
    \item The superposition between `0' and `1' enables to encode N bit information in $log_2{N}$ qubits.
    \item The reduction in features space also means reduction in number of layers and nodes needed. 
\end{itemize}

However these advantages come with their caveats. Even though the notion of the quantum parallelism is theoretically sound, the ``quantum advantage'' by which a quantum algorithm outperforms a classical one has not yet been experimentally demonstrated. This is because it is extremely difficult to maintain quantum phenomena in real physical systems \cite{nielsenandchuang}. The present quantum systems are highly susceptible to external noise resulting in depletion of the superposition. This phenomenon is called \textit{decoherence}. This issue is usually addressed by a phenomenon called error-correction, though decoherence must be suppressed below a certain threshold to achieve fault-tolerant operation and the physical resource demands of current error-correcting codes are significant.

\subsubsection{Encoding Input in a Quantum Register}

Indeed, the state of N qubits can be mathematically represented using $2^N$ dimensional Hilbert Space. We encode our $2^\frac{N}{2}$ by $2^\frac{N}{2}$  bit input image in this Hilbert space.  Let us consider $2^N$ bits of classical input ($\vec{x}$) to be encoded in a quantum state $\ket{\psi(x)}$ with $N$ qubits. Encoding the input in the coefficients of basis states is shown below: 
\begin{equation*}
    \ket{\psi(x)} = \sum_{j=1}^{2^N} x_j \ket{j}
\end{equation*}

\subsection{Quantum Neural Network based on Time Dependent Schrödinger Equation}

Although the Variational Quantum Circuit model fairly mimics a quantum analogue to a classical neural network, they are not theoretically congruent. In other words, a fully connected neural network with multiple layers cannot be exactly converted to a Variational Quantum Circuit. Another important factor would be to focus on using unique quantum properties like entanglement as most of these quantum operations can be simulated in a classical system using supercomputers. Hence, we propose a quantum neural network model based on these important facts. Our main motivation behind modelling this quantum neural network is to utilize the quantum properties to the fullest.
\subsection{The Proposed Model}
Consider a quantum register with an initial state $\ket{\psi}(0)$ with an input encoded. The time evolution of any Quantum State can be defined as follows,
\begin{equation} \label{eqn:QS}
    \ket{\psi(t)} = e^{-i \hat{H} t} \ket{\psi(0)}.
\end{equation}
where, $\hat{H}$ is called the Hamiltonian of the system which dictates the evolution of the system. Hence, we will define the Evolution Structure of Quantum Neural Network in the Hamiltonian basis. 

\begin{equation}
    \hat{H} = \hat{\sum\limits_{j=1}^{2^N} w_j \Xi_j}.
\end{equation}
where $w_j$ are real numbers defined as trainable weights.  $\Xi_j$ denotes any fundamental quantum gates like the pauliX and Identity gates. We denote the operation in Eqn. \ref{eqn:QS} as $\hat{\mathcal{N}}$. If we insert this Hamiltonian into Eqn. \ref{eqn:QS}, we get,
\begin{equation}
     \begin{aligned}
     \hat{\mathcal{N}} \ket{\psi} = \, &e^{-i\hat{H}\delta t}\ket{\psi} \\
     =\,  &e^{-i\hat{\sum\limits_{j=1}^{2^N} w_j \Xi_j}\delta t}\ket{\psi}\\
     =\,  &\prod_{j=1}^{2^N} e^{ - iw_j \Xi_j \delta t} \ket{\psi}\\
     =\,  &e^{ - iw_1 \Xi_1 \delta t} * e^{ - iw_2 \Xi_2 \delta t} \dots e^{ - iw_k \Xi_k  \delta t} \dots e^{ - iw_{2^N} \Xi_{2^N}  \delta t}\ket{\psi}. \\
\end{aligned}
\end{equation}

There is no requirement to introduce any logistic non-linear function since this system is intrinsically non-linear. Eqn. (8) is the theoretical definition of our proposed quantum neural network. We can translate any of these matrix exponentials to physical quantum gate matrices (Table 1) \cite{arashthesis}. Consider a matrix exponential as shown below for a one-qubit case with pauliX operator and weight w: 
\begin{equation*}
    \begin{aligned}
        e^{ - i w \sigma_x \delta t} &= cos(w \delta t)I - i sin(w \delta t) \sigma_x \\ 
        &= [\begin{smallmatrix} cos(w \delta t) &0 \\ 0 & cos(w \delta t) \end{smallmatrix}] + [\begin{smallmatrix} 0 &-i sin(w \delta t) \\ -i sin(w \delta t) & 0 \end{smallmatrix}] \\
        &= [\begin{smallmatrix} cos(w \delta t) &-i sin(w \delta t) \\ -i sin(w \delta t) & cos(w \delta t) \end{smallmatrix}] 
    \end{aligned}
    \label{1}
\end{equation*}

Hence, we can construct parameterized quantum gates for any variational quantum circuit with this definition. This can be achieved to model any kind of rotation operation in any direction as long as the matrices defined in $\Xi$ are unitary. 


\subsubsection{Defining Loss and Training}
\label{subsec:loss}
To optimize our QNN, we need to define a loss function to measure the performance of the network. We first need to encode the labels from the dataset in quantum states (say $\ket{y}$). Since we are working with complex numbers, squaring a complex number means multiplying with its conjugate. If the dataset size is m, we define the cost function as shown below:

\begin{equation}
\begin{aligned}
    \mathcal{L} &= (\bra{ \psi}  \hat{\mathcal{N}}^{\star} - \bra{y})*(\hat{\mathcal{N}}\ket{\psi}-\ket{y})\\
    &=\bra{ \psi}  \hat{\mathcal{N}}^{\star} \hat{\mathcal{N}}\ket{\psi} - \bra{y}\hat{\mathcal{N}}\ket{\psi} - \bra{\psi}\hat{\mathcal{N}}^*\ket{y}+\langle y | y \rangle\\
    &= 2 -\bra{y}\hat{\mathcal{N}}\ket{\psi} - \bra{\psi}\hat{\mathcal{N}}^*\ket{y}
\end{aligned}
\end{equation}
Where $\bra{\psi}$ and $\hat{\mathcal{N}}^*$ refer to the complex conjugates of $\ket{\psi}$ and  $\hat{\mathcal{N}}$ respectively. To optimize the neural network we need to minimize this cost function. This can be done by computing the gradient of the cost with respect to the weights and updating the weights to optimize the network as given below. 

\begin{equation*}
\begin{aligned}
    W &= W - \eta \nabla \mathcal{L}; \, \, \, \, \, \, \,   
    \nabla \mathcal{L} = [\frac{\partial \mathcal{L}}{\partial w _1}, \frac{\partial \mathcal{L}}{\partial w _2}, \frac{\partial \mathcal{L}}{\partial w _3} ........ \frac{\partial \mathcal{L}}{\partial w _n}]
\end{aligned}
\end{equation*}

We would like to compute the partial with respect to $k^{th}$ weight. This process is shown below:

\begin{equation}
\begin{aligned}
    \frac{\partial  \mathcal{L}}{\partial w_k} &= -\bra{y}\frac{\partial \hat{\mathcal{N}}}{\partial w_k}\ket{\psi} - \bra{\psi}\frac{\partial \hat{\mathcal{N}}^*}{\partial w_k}\ket{y}; \\
    \frac{\partial \hat{\mathcal{N}}}{\partial w_k} &= e^{ - i\alpha_1 \Xi_1 \delta t} * e^{ - i\alpha_2 \Xi_2 \delta t} \dots (-i \Xi_k \delta t) \,e^{ - i\alpha_k \Xi_k  \delta t} \dots e^{ - i\alpha_n \Xi_n  \delta t};\\
     \frac{\partial \hat{\mathcal{N}}^*}{\partial w_k} &= e^{  i\alpha_1 \Xi_1 \delta t} * e^{ i\alpha_2 \Xi_2 \delta t} \dots (i \Xi_k \delta t) \,e^{  i\alpha_k \Xi_k  \delta t} \dots e^{  i\alpha_n \Xi_n  \delta t}.
\end{aligned}
\label{qbackprop}
\end{equation}

The derivatives in equation \ref{qbackprop} can also be computed numerically using finite difference method as shown in \cite{killioran}. 
\section{Experiments}

A fundamental model of QNNs has been defined in previous Section. In this section, we show how these can be converted to specific circuit models. The problem we tackled is classification of the MNIST handwritten digit database. 

\subsection{Variational Quantum Circuits on MNIST Dataset}

\begin{figure}
    \centering
    \includegraphics[scale=0.5]{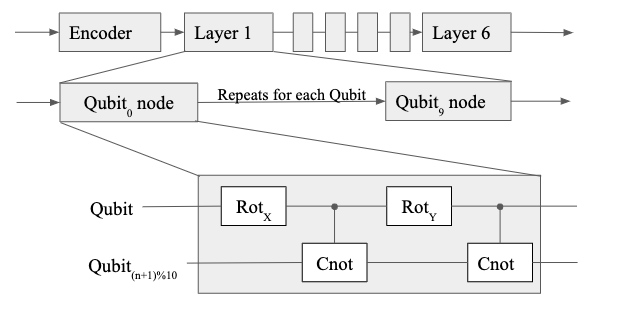}
    \caption{Schematic of the Variational Quantum circuit used for this case. Every layer is defined as a series of rotation gates in the x and y directions. Every qubit is linked to one other using CNOT gates. }
    \label{fig:qnnschematic}
\end{figure}

The MNIST dataset contains 60,000 training samples and 10,000 testing samples. Each of the $28\times28$ grey scale images contains a single handwritten digit that the model must correctly identify. 
In order to encode $28\times28$ values in a quantum state, the model requires 10 qubits. So a high performance quantum circuit simulation library called QuEST has been used in this experiment. The only prepossessing that wasperformed on the images was to zero pad them to $32 \times 32$ to fill the $2^{n}$ quantum space. To model the quantum circuit, we arbitrarily choose a structure and fine tuned it based on the performance. This is similar to the notion of deciding the number of layers and nodes for classical ANNs. Therefore, the structure was defined in a completely in an ad-hoc fashion. The circuit consists of the image encoder followed by the modular structure shown in Figure \ref{fig:qnnschematic}. On a quantum computer, these probabilities could be estimated by running the circuit many times and forming a probability histogram. In our simulations, we measured these probabilities directly. We used the mean square error between the normalized probability and a one-hot representation of the label to calculate the loss as shown in Section \ref{subsec:loss}. The weights were updated using mini-batch gradient descent, with a batch size of 10 images. 

\subsection {Results and Discussion}

\begin{figure}[h]
     \centering
      \begin{subfigure}[b]{0.5\textwidth}
         \centering
         \includegraphics[scale=0.175]{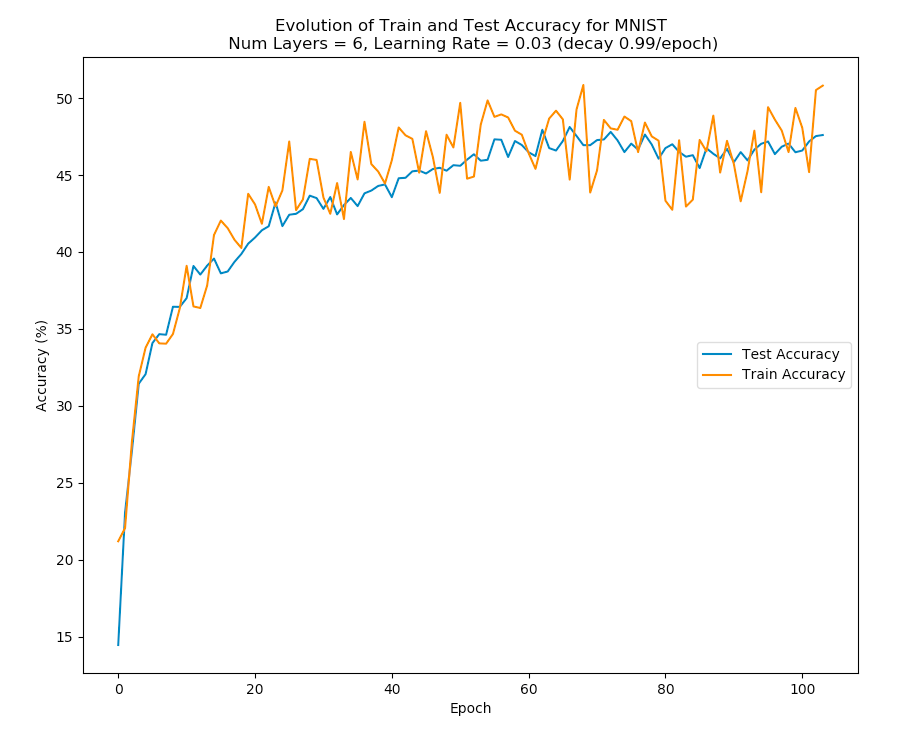}
         \caption{}
         \label{fig:10 layers}
     \end{subfigure}
     \hfill
     \begin{subfigure}[b]{0.5\textwidth}
         \centering
         \includegraphics[scale=0.175]{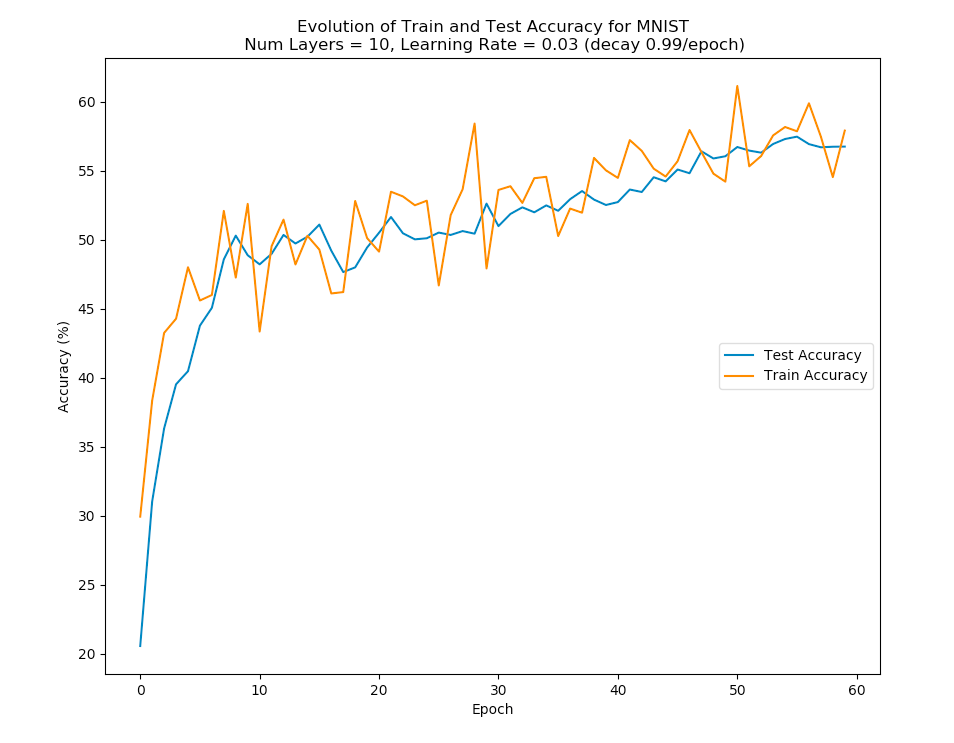}
         \caption{}
         \label{fig:6 layers}
     \end{subfigure}

        \caption{Visualizing the Training Process in relation to number of quantum layers. We see increased accuracy with number of layers.}
        \label{results}
\end{figure}

The proposed method experiments a number of trials based on number of quantum layers. The proposed method achieves 64\% of the recognition accuracy, the highest among the current quantum neural network models. 
In this experiment, the learning rate set as $\rho = 0.03$ which decays by 0.99 every epoch. We see an increase in performance with increase in number of layers. The experiment results are laid out in Table \ref{tab:resulttable}.  In Figure \ref{results}, the optimization process can be visualized as the evolution of network accuracy with every epoch. However, this was run in a simulation rather a real quantum system since there are very few accessible quantum computers with 10 usable qubits. This is an important point because simulating exponentially increasing dimensional Hilbert spaces in classical computers requires enormous amounts of resources. However, in an actual physical quantum computer most of the matrix transformations are computed naturally and require no computational power, potentially making QNN's much superior to classical ANN's in terms of performance.  

\begin{table}[h]
    \centering
    \begin{tabular}{|c|c|c|c|}
        \hline
         Number of Layers & Train Accuracy (\%) &  Test Accuracy (\%) & Convergence in (Epochs) \\
         \hline
         4 & 38.8 & 37.3 & 92\\  
         \hline
         6 & 47.0 & 50.1 & 103\\  
         \hline
         10 & 56.7 & 57.2 & 59\\  
         \hline
         \textbf{20} & \textbf{64.08} & \textbf{64.74} & \textbf{10}\\  
         \hline
    \end{tabular}
    \caption{Experimental Results.}
    \label{tab:resulttable}
\end{table}

\section{Conclusion and Future Work}

In this paper, we have presented an algorithm for defining and training Neural Networks in the perspective of time evolution of quantum states. This model is fundamental and hence can be inherited by any variants of quantum computing models. In addition, we derived a quantum backpropagation algorithm to train the QNN model and validate this algorithm for the MNIST dataset on a quantum computer simulation. The future work will be focused on addressing the Information Loss of Quantum States due to measurement. This problem can be addressed by maximally entangling the quantum state. Our next steps will be conducted in running and testing the QNN model in short term Quantum Processors.

\bibliographystyle{unsrt}
\bibliography{bibliography}

\end{document}